\documentclass[10.5pt,a4paper]{article}
\usepackage[T1]{fontenc}
\usepackage{lmodern}
\usepackage{geometry}
\usepackage{setspace}
\usepackage{graphicx}
\usepackage{booktabs}
\usepackage{amsmath, amssymb, amsthm}
\usepackage{hyperref}
\usepackage{caption}
\usepackage{subcaption}
\usepackage{color}
\usepackage{multirow}
\usepackage{orcidlink}
\hypersetup{
  colorlinks = true,
  linkcolor  = blue,
  citecolor  = blue,
  urlcolor   = blue
}

\title{Informative missingness and its implications in semi-supervised learning}

\author{
  Jinran Wu\orcidlink{0000-0002-2388-3614}$^{1}$,
  You-Gan Wang\orcidlink{0000-0003-0901-4671}$^{2,3}$,
  Geoffrey J.~McLachlan\orcidlink{0000-0002-5921-3145}$^{1}$\thanks{Corresponding author: g.mclachlan@uq.edu.au}
}
\date{
  \vspace{1em}
  $^{1}$School of Mathematics and Physics, The University of Queensland, Australia\\
  $^{2}$School of Statistics and Data Science, Guangdong University of Finance \& Economics, China\\
  $^{3}$School of Mathematics and Computational Science/Hunan Key Laboratory for Computation and Simulation in Science and Engineering, Xiangtan University, China
}

\begin{document}

\maketitle

\begin{abstract}
Semi-supervised learning (SSL) constructs classifiers using both labelled and unlabelled data. It leverages information from labelled samples, whose acquisition is often costly or labour-intensive, together with unlabelled data to enhance prediction performance. This defines an incomplete-data problem, which statistically can be formulated within the likelihood framework for finite mixture models that can be fitted using the expectation-maximisation (EM) algorithm. Ideally, one would prefer a completely labelled sample, as one would anticipate that a labelled observation provides more information than an unlabelled one. However, when the mechanism governing label absence depends on the observed features or the class labels or both, the missingness indicators themselves contain useful information. In certain situations, the information gained from modelling the missing-label mechanism can even outweigh the loss due to missing labels, yielding a classifier with a smaller expected error than one based on a completely labelled sample analysed~\cite{AhfockMcLachlan2020}. This improvement arises particularly when class overlap is moderate, labelled data are sparse, and the missingness is informative. Modelling such informative missingness thus offers a coherent statistical framework that unifies likelihood-based inference with the behaviour of empirical SSL methods.
\end{abstract}

\noindent
In many classification tasks, only a subset of class labels is typically observed. In practice, labels are often omitted for ambiguous instances or are recorded preferentially by class. In the terminology of missing-data mechanisms~\cite{Rubin1976}, such missingness is informative. A likelihood-based formulation allows this informative mechanism to be modelled explicitly. The joint distribution of features and class labels can be represented by a finite mixture model, with estimation performed via EM~\cite{Dempster1977}. The pattern of missing labels itself conveys information about where classes overlap and how labels are allocated. Recent analyses have shown that partially labelled data analysed with a correctly specified mechanism can produce more accurate predictions than completely labelled data~\cite{AhfockMcLachlan2020,ahfock2023semi}. The focus here is on likelihood structures rather than empirical construction. The relationship between missingness mechanisms and their implications for semi-supervised learning (SSL) is summarised schematically in Figure~\ref{fig:overview}.

\begin{figure}[!ht]
  \centering
  \includegraphics[width=0.75\textwidth]{TII_figure.png}
  \caption{Schematic illustration showing how MCAR, MAR, and MNAR missingness mechanisms influence model-based inference and classification boundaries in SSL. When the missingness mechanism is correctly modelled within a finite-mixture framework, a classifier trained on partially labelled data can outperform one trained on completely labelled data~\cite{AhfockMcLachlan2020}.}
  \label{fig:overview}
\end{figure}

\section*{Likelihood Formulation for Partially Labelled Data}

Let $(\mathbf{Y}, Z, M)$ denote the random variables corresponding to the feature vector, class label, and missingness indicator, respectively, where $\mathbf{Y}\in\mathbb{R}^p$, $Z\in\{1,\ldots,g\}$, and $M\in\{0,1\}$ indicates whether the label is missing ($M=1$) or observed ($M=0$). Assume that $\mathbf{Y}$ follows a finite mixture model with $g$ components, $f(\mathbf{y};\boldsymbol{\theta})=\sum_{i=1}^{g}\pi_i f_i(\mathbf{y};\boldsymbol{\omega}_i),$ where $\pi_i (>0)$ is the class proportion and $\sum_{i=1}^{g}\pi_i=1$, $f_i(\cdot;\boldsymbol{\omega}_i)$ is the class-conditional density, and $\boldsymbol{\theta}=\{\pi_i,\boldsymbol{\omega}_i\}_{i=1}^{g}$ is the full parameter set. Given data $\{(\mathbf{y}_j,z_j,m_j):j=1,\ldots,n\}$, the observed-data log-likelihood can be written as
\vspace{-4mm}
\begin{equation}
\label{eq:loglik}
\log L(\boldsymbol{\theta})
= \sum_{j=1}^{n} (1 - m_j)\sum_{i=1}^{g}z_{ij}\log\{\pi_{i} f_{i}(\mathbf{y}_j; \boldsymbol{\omega}_{i})\}
+ \sum_{j=1}^{n} m_j\log\!\Bigl\{\sum_{i=1}^{g} \pi_i f_i(\mathbf{y}_j; \boldsymbol{\omega}_i)\Bigr\},
\end{equation}
\vspace{-4mm}
\newline
where $z_{ij}=1$ if sample $j$ belongs to class $i$, and $0$ otherwise. The first term corresponds to labelled observations ($M=0$), while the second integrates over unlabelled cases ($M=1$). When all labels are observed, \eqref{eq:loglik} reduces to the likelihood in the completely labelled situation; where all the labels are missing, it reduces to the standard mixture model-based clustering form. Thus, partially labelled data can be viewed as an intermediate setting between these extremes, where both labelled and unlabelled observations contribute to parameter estimation within a joint likelihood framework. It shall be noted that the posterior class probabilities $\tau_{ij}$ for unlabelled observations $\mathbf{y}_j$ provide fractional estimates of cluster memberships. This likelihood-based formulation is implemented within the EM framework for SSL~\cite{Dempster1977} by which inference with partially labelled data is a natural instance of incomplete-data estimation~\cite{McLachlanLeeRathnayake2019}.

\section*{Informative Missingness: Mechanism and Estimation}

In the past, the focus has been on the theoretical results for situations where class missingness is ignorable. In many applications, however, the probability that a label is missing may depend on the data itself and therefore be informative. Two common forms of informative missingness are considered~\cite{Rubin1976}. Under missing at random (MAR), the probability of a label being unobserved depends on the observed feature vector $\mathbf{y}_j$ through a measure of classification uncertainty, such as entropy $e(\mathbf{y}_j;\boldsymbol{\theta})=-\sum_{i=1}^{g}\tau_{ij}\log \tau_{ij}$. A convenient specification adopts a logistic form, $\operatorname{logit}\{\Pr(M_j=1\mid \mathbf{Y}=\mathbf{y}_j)\}=\boldsymbol{\xi}^\top \mathbf{u}(\mathbf{y}_j)$, where $\mathbf{u}(\mathbf{y}_j)$ may include $e(\mathbf{y}_j;\boldsymbol{\theta})$ or related features, and $\boldsymbol{\xi}$ denotes the mechanism parameters. This captures feature-dependent missingness: ambiguous instances near class boundaries are less likely to be labelled. Under missing not at random (MNAR), the probability of being unlabelled also depends on the true class label and feature vectors, $\Pr(M_j=1\mid Z=i,\mathbf{Y}=\mathbf{y}_j)=\phi_{i}(b_{\mathbf{y}_j})$, where $\phi_{i}$ represent the class-dependent missingness probability evaluted at the discriminant term $b_{\mathbf{y}_j}$ for sample $\mathbf{y}_j$ from class $i$. This reflects class-dependent or preferential missingness, where annotation effort or accessibility varies systematically across classes. Both MAR and MNAR correspond to non-ignorable mechanisms in the sense of Rubin~\cite{Rubin1976}, and the indicators $M_j$ thereby contain additional information about decision boundaries and class proportions.

\medskip
\noindent
From an information-theoretic perspective, the total Fisher information (the expected curvature of the log-likelihood) in a partially labelled sample can be decomposed into three components~\cite{AhfockMcLachlan2020}. Let $\mathbf{I}_{\mathrm{CC}}$ denote the information that would be available under complete classification, $\mathbf{I}_{\mathrm{UC}}$  the information in the unlabelled data, and let $\mathbf{I}_{\mathrm{CC}}^{(\mathrm{clr})}$ represent the conditional information for logistic regression~\cite{ahfock2023semi}. The contribution of the missingness mechanism itself is captured by $\mathbf{I}_{\mathrm{PC}}^{(\mathrm{miss})}$, which quantifies the information gained by modelling how labels are missing. The total information can be written schematically as
\vspace{-3mm}
\begin{equation*}
\begin{aligned}
\mathbf{I}_{\mathrm{PC}}^{(\mathrm{full})}
&= (1 - \gamma)\,\mathbf{I}_{\mathrm{CC}}
+ \gamma\,\mathbf{I}_{\mathrm{UC}} 
+ \mathbf{I}_{\mathrm{PC}}^{(\mathrm{miss})} \\
&= (1 - \gamma)\,\mathbf{I}_{\mathrm{CC}}
+ \gamma\,(\mathbf{I}_{\mathrm{CC}}
- \mathbf{I}_{\mathrm{CC}}^{(\mathrm{clr})})
+ \mathbf{I}_{\mathrm{PC}}^{(\mathrm{miss})} \\
&= \mathbf{I}_{\mathrm{CC}}
- \gamma\,\mathbf{I}_{\mathrm{CC}}^{(\mathrm{clr})}
+ \mathbf{I}_{\mathrm{PC}}^{(\mathrm{miss})},
\end{aligned}
\end{equation*}
\vspace{-2mm}
\newline
where $\gamma$ is the expected proportion of missing labels. The first term measures the potential information under full supervision; the second subtracts the loss due to incomplete labels; and the third adds the gain from modelling the missingness mechanism. When the mechanism is non-informative, $\mathbf{I}_{\mathrm{PC}}^{(\mathrm{miss})}=0$, and the total information decreases linearly with $\gamma$. However, when the missingness depends on features or latent class structure, $\mathbf{I}_{\mathrm{PC}}^{(\mathrm{miss})}>0$ and can exceed $\gamma\,\mathbf{I}_{\mathrm{CC}}^{(\mathrm{clr})}$, producing the apparent paradox that partially labelled data analysed under a correctly specified mechanism may be more informative than completely labelled data analysed. In such cases, the missing-label indicators effectively serve as auxiliary variables, sharpening inference on the class boundaries and mixing proportions. The magnitude of this gain is highest where class overlap is moderate, the labelled subset is small, and the missingness mechanism is strongly dependent on feature uncertainty or class membership. Intuitively, this means that when missingness carries useful structure, the absence of some labels can itself provide additional information for learning.

\medskip
\noindent
The likelihood approach provides a direct way to formulate models with informative missingness. The observed-data log-likelihood is extended to include the contribution of the missingness indicators ${M_j}$, so that both the mixture and mechanism parameters are estimated jointly, typically via the EM algorithm. This allows uncertainty to be propagated coherently and yields efficient estimates when the mechanism is specified. Gains are most visible when class overlap is moderate and labelled data are sparse.

\section*{Relation to Empirical SSL}

Most SSL algorithms implicitly assume how labels are missing and how uncertainty determines which unlabelled data contribute to learning. Classical SSL theory rests on the smoothness and low-density assumptions, suggesting that decision boundaries should lie in regions of high uncertainty or low sample density. Under this view, minimum-entropy regularisation~\cite{GrandvaletBengio2004} corresponds to an MAR mechanism, where unlabelled data concentrate near ambiguous regions. This interpretation extends naturally to deep learning methods such as MixMatch~\cite{Sohn2020}, which combines guessed labels and data augmentation to enforce prediction consistency, thereby reducing entropy in uncertain regions. From this standpoint, these approaches can be regarded as empirical realisations of MAR-type mechanisms that regulate the influence of unlabelled data during training-implicitly recognising that the absence of labels is itself informative.

\medskip
\noindent
Recent surveys have systematically mapped the evolution of SSL from probabilistic mixture formulations to deep neural frameworks~\cite{yang2022survey,gui2024survey}. As summarised by Yang et~al.~\cite{yang2022survey}, empirical SSL methods fall into five major paradigms: deep generative models, consistency regularisation, graph-based inference, pseudo-labelling, and hybrid frameworks that integrate multiple strategies for state-of-the-art performance. Despite architectural diversity, these paradigms share a common inductive bias: unlabelled data are not treated as MCAR but are weighted by their predictive uncertainty, consistency, or neighbourhood structure conceptually similar to an entropy-driven MAR mechanism. Gui et~al.~\cite{gui2024survey} further showed that class imbalance induces systematic bias in pseudo-labelling and consistency-based models, mirroring class-dependent or preferential missingness analogous to an MNAR mechanism. These observations highlight that the performance of deep SSL depends critically on how the implicit label-missingness process is represented. Hence, empirical SSL algorithms can be interpreted as heuristic approximations to informative missingness, where entropy minimisation and distribution alignment serve as empirical surrogates for the statistical mechanisms governing label absence. This perspective provides a conceptual bridge between likelihood-based inference and deep learning practice, offering a foundation for unifying theoretical and empirical SSL within a mechanism-aware framework.

\section*{Concluding Remarks}

The central objective is to utilise informative missingness to enhance the predictive performance of classifiers trained on partially labelled data. Missingness arising from regions of high uncertainty within the feature space can provide additional information that enhances model estimation and classification. Within a finite-mixture and EM framework, incorporating a MAR or MNAR mechanism enables the systematic exploitation of this information through a joint-likelihood formulation. The potential benefit is most pronounced in problems with moderate class overlap and limited labelled data. When the mechanism is well specified, the resulting classifier can achieve a smaller expected error than one trained on a completely labelled sample~\cite{AhfockMcLachlan2020, ahfock2023semi}. Modelling of the missing-label mechanism is therefore both practical and important for achieving reliable and accurate inference.

\medskip
\noindent
Several avenues for future exploration remain open. From a theoretical perspective, further work should aim to establish general information-theoretic bounds that quantify how missingness-dependent uncertainty contributes to Fisher information and generalisation performance, and to formalise the link between entropy-based MAR mechanisms and empirical SSL objectives. Additionally, its sensitivity to model misspecification and robustness within the Godambe information bound merit future investigation. At the methodological level, developing hybrid algorithms that learn or approximate the missingness mechanism, for example, entropy-driven weighting schemes or neural architectures that jointly estimate $\Pr(M \mid \mathbf{Y})$ alongside deep neural network parameters, could further integrate likelihood-based inference with deep SSL. In terms of applications, informative-missingness modelling can be extended to domains such as medical diagnosis, environmental sensing, and geoscientific monitoring, where label incompleteness often reflects systematic or human-driven bias rather than randomness. Embedding informative missingness within the broader SSL paradigm provides a coherent pathway to unify classical statistical modelling and deep learning practice, advancing the theoretical, methodological, and applied frontiers of data-efficient learning.

\section*{Acknowledgments}
This work was supported by the Australian Research Council [DP230101671], ARC Training Centre on Innovation in Biomedical Imaging Technology [IC170100035], and the Science and Technology Innovation Program of Hunan Province [2022RC4028].

\section*{Declaration of interests}
The authors declare no competing interests.

\end{document}